Please cite this article as:

Yukun Bao, Tao Xiong, Zhongyi Hu, "Multi-Step-Ahead Time Series Prediction using Multiple-Output Support Vector Regression."2013, accepted, *Neurocomputing*.

# Research Highlights

➢ Propose M-SVR model with MIMO strategy for multi-step-ahead time series prediction.

➢ Provide empirical evidence on three multi-step-ahead prediction strategies using SVR.

➢ The M-SVR using MIMO strategy is the best with accredited computational load.

➢ The computational load of standard SVR using direct strategy is extremely expensive.

➢ The standard SVR using iterated strategy is best in terms of low computational load.

# Multi-Step-Ahead Time Series Prediction using Multiple-Output

# Support Vector Regression

Yukun Bao[*], Tao Xiong, Zhongyi Hu

Department of Management Science and Information Systems,

School of Management, Huazhong University of Science and Technology, Wuhan, P.R.China,

430074

## Abstract

Accurate time series prediction over long future horizons is challenging and of

great interest to both practitioners and academics. As a well-known intelligent

algorithm, the standard formulation of Support Vector Regression (SVR) could be

taken for multi-step-ahead time series prediction, only relying either on iterated

strategy or direct strategy. This study proposes a novel multiple-step-ahead time series

prediction approach which employs multiple-output support vector regression

(M-SVR) with multiple-input multiple-output (MIMO) prediction strategy. In addition,

the rank of three leading prediction strategies with SVR is comparatively examined,

providing practical implications on the selection of the prediction strategy for

multi-step-ahead forecasting while taking SVR as modeling technique. The proposed

approach is validated with the simulated and real datasets. The quantitative and

comprehensive assessments are performed on the basis of the prediction accuracy and

computational cost. The results indicate that: 1) the M-SVR using MIMO strategy

achieves the best accurate forecasts with accredited computational load, 2) the

standard SVR using direct strategy achieves the second best accurate forecasts, but

with the most expensive computational cost, and 3) the standard SVR using iterated

[*] Corresponding author: Tel: +86-27-87558579; fax: +86-27-87556437.
Email: yukunbao@hust.edu.cn or y.bao@ieee.org

strategy is the worst in terms of prediction accuracy, but with the least computational cost.

**Keywords:** Multi-step-ahead time series prediction; Multiple-input multiple-output (MIMO) strategy; Multiple-output support vector regression (M-SVR)

# 1. Introduction

Time series prediction in general, and multi-step-ahead time series prediction in particular, has been the focus of research in many domains. In one-step-ahead prediction, the predictor uses all or some of the observations to estimate a variable of interest for the time-step immediately following the latest observation. Predicting two or more steps ahead, named as multi-step-ahead prediction, however, appeal not only to researchers, but also, probably more importantly, to policy makers, stockbrokers and other practitioners. Nevertheless, unlike one-step-ahead prediction, multi-step-ahead prediction faces typically growing amount of uncertainties arising from various sources. For instance, an accumulation of errors and lack of information make multi-step-ahead prediction more difficult [1]. Thus, given a specific modeling technique, selecting suitable modeling strategy for multi-step-ahead time series prediction has been a major research topic that has significantly practical implication.

At present, there are two commonly used modeling strategies, namely iterated strategy and direct strategy, to generate multi-step-ahead forecasts. Iterated strategy constructs a prediction model by means of minimizing the squares of the in-sample one-step-ahead residuals, and then uses the predicted values as an input to predict the following forecast. Since it uses the predicted values from the past, it can be shown to be susceptible to the error accumulation problem [2, 3]. In contrast to the iterated strategy which constructs a single model, direct strategy first suggested by Cox [4] constructs a set of prediction models for each horizon using only its past observations, where the associated squared multi-step-ahead errors, instead of the squared one-step-ahead errors, are minimized [5]. The accumulation of errors in iterated strategy drastically deteriorates the prediction accuracy, while direct strategy is time consuming. Addressing these problems, Bontempi [6] introduced a multiple-input multiple-output (MIMO) strategy for multi-step-ahead prediction with the goal of preserving, among the predicted values, the stochastic dependency characterizing the time series, which facilitates to model the underlying dynamics of the time series. In this case, the predicted value is not a scalar quantity but a vector of future values whose size equals to the prediction horizon. Recently, experimental assessment of three aforementioned strategies, such as iterated strategy, direct strategy and MIMO strategy, on the NN3 and NN5 competition data showed that the MIMO strategy is very promising and able to outperform the two counterparts [7, 8]. It should be noted that the implementation modeling technique in [7, 8] is lazy learning, a local modeling technique with flexibility in constructing input-output modeling structure.

As a well-known intelligent algorithm, support vector machines (SVMs) have attracted particular attention from both practitioners and academics in terms of time series prediction (in the formulation of support vector regression (SVR)) during the last decade. They are found to be a viable contender among various time series modeling techniques (see [9-11]), and have been successfully applied to different areas (see [12-14]). Despite the promising MIMO strategy justified in [7, 8], the standard formulation of SVR cannot take the straightforward use of MIMO strategy for multi-step-ahead prediction due to its inherent single-output structure. Consequently studies highlighting the superiority of SVR for multi-step-ahead time series prediction have to rely either on iterated strategy [10, 15], direct strategy [16, 17], and recently a wise variant of direct strategy using a set of single output SVR models cooperated in an iterated way for multi-step-ahead prediction [18]. To generalize the SVR from regression estimation and function approximation to multi-dimensional problems, Pérez-Cruz et al. [19] proposed a multi-dimensional SVR that uses a cost function with a hyperspherical intensive zone, capable of obtaining better predictions than using an SVR independently for each dimension. Subsequently, Sanchez-Fernandez et al. [20] used the new SVR algorithm to deal with the problem of multiple-input multiple-output frequency nonselective channel estimation. More recently, Tuia et al. [21] proposed a multiple-output SVR model (M-SVR), based on the previous contribution in [20], for the simultaneous estimation of different biophysical parameters from remote sensing images. Upon the work of [19-21], M-SVR has been established and justified in a variety of disciplines, including communication [20], energy market [22], machine learning [23].

Although past studies have clarified the capability of M-SVR, there has been very few, if any, effort to examine the potential of M-SVR for multi-step-ahead time series prediction. As such, this study proposes a MIMO strategy based M-SVR approach for multi-step-ahead time series prediction, and then comparatively examines the rank of three leading prediction strategies by providing the first empirical evidence as whether the M-SVR is promising for generating accurate multi-step-ahead prediction, and which prediction strategy should be preferred in practice when using SVR as modeling technique. It should be noted that, although the comparative study of the three prediction strategies mentioned above have been conducted in [7, 8], the modeling technique used in these two published work is lazy learning. The applications of computational intelligence methods, e.g. SVR, with multiple inputs multiple outputs structure for multi-step-ahead time series prediction, had not yet been fully explored, which may shed a different light on the theoretical modeling issues and provide implications for practitioners. In addition, not only the prediction accuracy, but also the computational cost of the three aforementioned strategies is examined in the current study, highlighting the potential in massive computing environment. As such, the three examined models turned out to be: standard support vector regression using iterated strategy (abbreviated to ITER-SVR), standard support vector regression using direct strategy (abbreviated to DIR-SVR), and multiple-output support vector regression using MIMO strategy (abbreviated to MIMO-SVR). In addition, Naïve and Seasonal Naïve models are selected as benchmarks here because they are normally used to generate baseline forecasts. Both simulated (i.e., Hénon and Mackey-Glass time series) and real (i.e., NN3 competition data) datasets are adopted for the comparisons. The experimental results are judged on the basis of the prediction accuracy and computational cost.

This paper is structured as follows. In Section 2, we provide a brief introduction to the M-SVR and prediction strategies used in this study. Afterwards, Section 3 details the research design on data source, data preprocessing, accuracy measure, input selection, SVR implementation, and experimental procedure. Following that, in Section 4, the experimental results are discussed. Section 5 finally concludes this work.

## 2. Methodologies

## 2.1 M-SVR formulation

M-SVR proposed by Pérez-Cruz et al. [19] is a generalization of the standard SVR. Note that, though the M-SVR is well-studied in a variety of disciplines (see [20, 22, 23]), what is novel here is its application to multi-step-ahead time series prediction. Detailed discussions on the M-SVR can be found in [19-21], but a brief introduction about formulation is provided here.

Given a time series $\{\varphi_1, \varphi_2, \ldots, \varphi_N\}$, multi-step-ahead time series modeling and prediction is regarded as finding the mapping between the current and previous observation $\mathbf{x} = [\varphi_t, \varphi_{t-1}, \ldots, \varphi_{t-d+1}] \in \mathbb{R}^d$ and the future observation $\mathbf{y} = [\varphi_{t+1}, \varphi_{t+2}, \ldots, \varphi_{t+H}] \in \mathbb{R}^H$ from the training sample, i.e., $\{(\mathbf{x}_i, \mathbf{y}_i)\}_{i=d}^n$. The M-SVR solves this problem by finding the regressor $\mathbf{w}^j$ and $b^j$ $(j = 1, \ldots, H)$ for every output that minimizes:

$$L_p(\mathbf{W}, \ \mathbf{b}) = \frac{1}{2}\sum_{j=1}^{H}\left\|\mathbf{w}^j\right\|^2 + C\sum_{i=1}^{n}L(u_i) \qquad (1)$$

Where $u_i = \left\|\mathbf{e}_i\right\| = \sqrt{(\mathbf{e}_i^T \mathbf{e}_i)}$,

$\mathbf{e}_i^T = \mathbf{y}_i^T - \varphi(\mathbf{x}_i)\mathbf{W} - \mathbf{b}^T$

$\mathbf{W} = \left[\mathbf{w}^1, \ldots, \ \mathbf{w}^H\right]$,

$\mathbf{b} = \left[b^1, \ldots, \ b^H\right]^T$

$\varphi(\cdot)$ is an nonlinear transformation to the feature space which is higher dimensional space usually, and $C$ is a hyper parameter which determines the trade-off between the regularization and the error reduction term. $L(u)$ is a quadratic epsilon-insensitive cost function defined as Eq. (2), which is a differentiable form of the Vapnik $\varepsilon$ insensitive loss function.

$$L(u) = \begin{cases} 0 & u < \varepsilon \\ u^2 - 2u\varepsilon + \varepsilon^2 & u \geq \varepsilon \end{cases} \tag{2}$$

In Eq. (2) when $\varepsilon$ is nonzero, it will take into account all outputs to construct each individual regressor and will be to obtain more robust predications, then yield a single support vector set for all dimensions. It should be noted that the resolution of the proposed problem cannot be done straightforwardly, thus an iterative reweighted least squares (IRWLS) procedure based on quasi-Newton approach to obtain the desired solution was proposed by Sanchez-Fernandez et al. [20]. By introducing a first-order Taylor expansion of cost function $L(u)$, the objective of Eq. (1) will be approximated by the following equation

$$L_p'(\mathbf{W}, \ \mathbf{b}) = \frac{1}{2}\sum_{j=1}^{H}\left\|\mathbf{w}^j\right\|^2 + \frac{1}{2}\sum_{i=1}^{n}a_i u_i^2 + CT, \quad a_i = \begin{cases} 0 & u_i^k < \varepsilon \\ \dfrac{2C\left(u_i^k - \varepsilon\right)}{u_i^k} & u_i^k \geq \varepsilon \end{cases}$$

(3)

where $CT$ is constant term which does not depend on $\mathbf{W}$ and $\mathbf{b}$, and the superscript $k$ denotes $k$th iteration.

To optimize Eq. (3), an IRWLS procedure is constructed which linearly searched the next step solution along the descending direction based on the previous solution [20]. According to the Representer Theorem [24], the best solution of minimization of Eq. (3) in feature space can be expressed as $\mathbf{w}^j = \sum_i \beta_i^j \phi(\mathbf{x}_i)^{j} = \Phi^T \beta^j$, so the target of M-SVR is transformed into finding the best $\boldsymbol{\beta}$ and $\mathbf{b}$. The IRWLS of M-SVR can be summarized in the following steps [20, 22]:

Step 1: Initialization: Set $k = 0$, $\boldsymbol{\beta}^k = 0$ and $\mathbf{b}^k = 0$, calculate $u_i^k$ and $a_i$;

Step 2: Compute the solution $\boldsymbol{\beta}^s$ and $\mathbf{b}^s$ according to the next equation

$$\begin{bmatrix} \mathbf{K} + \mathbf{D}_a^{-1} & 1 \\ \mathbf{a}^T \mathbf{K}^a & 1^T \mathbf{a} \end{bmatrix} \begin{bmatrix} {}^j \\ b^j \end{bmatrix} = \begin{bmatrix} {}^j \\ \mathbf{a}^T \mathbf{y}^j \end{bmatrix}, \quad j = 1, 2, \ldots, H \tag{4}$$

where $\mathbf{a} = [a_1, \ldots, a_n]^T$, $(\mathbf{D}_a)_{ij} = a_i \delta(i - j)$, and $\mathbf{K}$ is the kernel matrix. Define the corresponding descending direction $\mathbf{p}^k = \begin{bmatrix} \mathbf{w}^s - \mathbf{w}^k \\ (\mathbf{b}^s - \mathbf{b}^k)^T \end{bmatrix}$.

Step 3: use a back tracking algorithm to compute $\boldsymbol{\beta}^{k+1}$ and $\mathbf{b}^{k+1}$, and further obtain $u_i^{k+1}$ and $a_i$. Go back to step 2 until convergence.

The convergence proof of the above algorithm is given in [20]. Because $u_i^k$ and $a_i$ are computed by means of every dimension of $\mathbf{y}$, each individual regressor contains the information of all outputs which can improve the prediction performance [22].

## 2.2 Strategies used in this study

Multi-step-ahead time series forecasting can be described as an estimation on future time series $\varphi_{N+h}, (h = 1, 2, \ldots, H)$, while $H$ is an integer and more than one, given the current and previous observation $\varphi_t, (t = 1, 2, \ldots, N)$. In the present study, iterated strategy, direct strategy, and MIMO strategy are selected for multi-step-ahead forecasting. For each selected strategies, there are a large number of variations proposed in the literatures, and it would be a hopeless task to consider all existing varieties. Our guideline was therefore to consider the basic version of each strategy (without the additions, or the modifications proposed by some other researchers). The reason for selecting the following three strategies is that they are some of the most commonly used strategies. The following subsection presents a detailed definition of the selected strategies.

1) Iterated strategy

The first is named as the iterated strategy by Chevillon [2] and is often advocated in standard time series textbooks (see [25, 26]). This strategy constructs a prediction model by means of minimizing the squares of the in-sample one-step-ahead residuals, and then uses the predicted value as an input for the same model when to forecast the subsequent point, and continues in this manner until reaching the horizon.

In more detail, iterated strategy first embeds the original series into an input-output format:

$$D = \left\{ \left( \mathbf{x}_t, y_t \right) \in \left( \mathbb{R}^m \times \mathbb{R} \right) \right\}_{t=d}^{N} \tag{5}$$

Where $\mathbf{x}_t \subset \left\{ \varphi_t, \ldots, \varphi_{t-d+1} \right\}$, $y_t = \varphi_{t+1}$.

Then the iterated prediction strategy learns one-step-ahead prediction model:

$$\varphi_{t+1} = f(\mathbf{x}_t) + \omega$$

where $\omega$ denotes the additive noise.

After the learning process, the estimation of the $H$ next values is returned by:

$$\hat{\varphi}_{t+h} = \begin{cases} \hat{f}\left( \varphi_t, \varphi_{t-1}, \ldots, \varphi_{t-d+1} \right) & if \quad h=1 \\ \hat{f}\left( \varphi_{t+h-1}, \ldots, \varphi_{t+1}, \varphi_t, \ldots, \varphi_{t-d+h} \right) & if \quad h \in \left[ 2, \ldots, d \right] \\ \hat{f}\left( \hat{\varphi}_{t+h-1}, \ldots, \varphi_{t+h-d} \right) & if \ h \in \left[ d+1, \ldots, H \right] \end{cases} \tag{6}$$

2) Direct strategy

In contrast to the iterated strategy which uses a single model, the other commonly applied strategy, namely direct strategy first suggested by Cox [4], constructs a set of prediction models for each horizon using only its past observations, where the associated squared multi-step-ahead errors are minimized [5]. Direct strategy estimates $H$ different models between the inputs and the $H$ outputs to predict $\{\varphi_{N+h}, h = 1, 2, \ldots, H\}$ respectively.

The direct strategy first embeds the original series into $H$ datasets

$$
\begin{aligned}
D_1 &= \left\{ (\mathbf{x}_t, y_{t1}) \in \left( \mathbb{R}^m \times \mathbb{R} \right) \right\}_{t=d}^{N}, \\
&\vdots \\
D_H &= \left\{ (\mathbf{x}_t, y_{tH}) \in \left( \mathbb{R}^m \times \mathbb{R} \right) \right\}_{t=d}^{N}.
\end{aligned}
\tag{7}
$$

where $\mathbf{x}_t \subset \{\varphi_t, \ldots, \varphi_{t-d+1}\}$, $y_{th} = \varphi_{t+h}$.

Then, the direct prediction strategy learns $H$ direct models on $D_h \in \{D_1, \ldots, D_H\}$, respectively.

$$
\varphi_{t+h} = f_h(\mathbf{x}_t) + \omega_h, h \in \{1, \ldots H\}.
\tag{8}
$$

Where $\omega$ denotes the additive noise.

After the learning process, the estimation of the $H$ next values is returned by:

$$
\hat{\varphi}_{t+h} = \hat{f}_h(\varphi_t, \varphi_{t-1}, \ldots, \varphi_{t-d+1}), \quad h \in \{1, \ldots, H\}.
\tag{9}
$$

3) MIMO strategy

The last strategy, namely MIMO, was first proposed by Bontempi [6] and characterized as an approach structured as multiple-input multiple-output, where the predicted value is not a scalar quantity but a vector of future values $(\varphi_{N+1}, \varphi_{N+2}, \ldots, \varphi_{N+H})$ of the time series $\varphi_t (t = 1, 2, \ldots, N)$. Compared with the direct strategy, which estimates $\varphi_{N+h} (h = 1, 2, \ldots, H)$ using $H$ models, MIMO employs only one multiple-output model, preserving the temporal stochastic dependency hidden in the predicted time series.

MIMO strategy first embeds the original series into datasets

$$D = \left\{ (\mathbf{x}_t, \mathbf{y}_t) \in \left( \mathbb{R}^m \times \mathbb{R}^H \right) \right\}_{t=d}^N . \tag{10}$$

where $\mathbf{x}_t \subset \{ \varphi_t, \ldots, \varphi_{t-d+1} \}, \mathbf{y}_t = \{ \varphi_{t+1}, \ldots, \varphi_{t+H} \}$

Then MIMO prediction strategy learns multiple-output prediction model

$$\mathbf{y}_t \, \boldsymbol{\omega} \, f(\mathbf{x}_t) + \tag{11}$$

After learning process, the estimation of the $H$ next values are returned by

$$\left\{ \widehat{\varphi_{t+1}}, \ldots, \varphi_{t+H} \right\} = \hat{f}(\varphi_t, \ldots, \varphi_{t-d+1}) \tag{12}$$

# 3. Research design

## 3.1 Data and preprocessing

To evaluate the performances of the proposed M-SVR using MIMO strategy and the counterparts in terms of the forecast accuracy, two simulated time series, i.e., Hénon and Mackey-Glass time series, and a real world dataset, i.e., NN3 competition dataset, are used in this present study.

Hénon and Mackey-Glass time series are recognized as benchmark time series that have been commonly used and reported by a number of studies related to time series modeling and forecasting [27-30].

The Hénon map is one of the most studied dynamic systems. The canonical Hénon map takes points in the plane following Hénon [31].

$$\begin{aligned} \phi_{t+1} &= \varphi_t + 1 - 1.4\phi_t^2 \\ \varphi_{t+1} &= 0.3\phi_t \end{aligned} \tag{13}$$

The Mackey-Glass time series is approximated from the differential Eq. (14) (see [32]).

$$\frac{d\varphi_t}{dt} = \frac{0.2\varphi_{t-\tau}}{1 + \varphi_{t-17}^{10}} - 0.1\varphi_t \tag{14}$$

For each data-generating process (DGP), i.e., Hénon and Mackey-Glass process, we simulate twenty time series with different initialization and sample size, as is shown in Table 1. The data for these time series are generated by the *Chaotic Systems Toolbox*[1] from the MATLAB software.

**<Insert Table 1 here>**

The NN3 competition was organized in 2007, targeting computational-intelligence forecasting approaches. The competition dataset of 111 monthly time series drawn from homogeneous population of real business time series is used for evaluation[2]. The data are of a monthly reference, with positive observations and structural characteristics which vary widely across the time series. For example, many of the series are dominated by a strong seasonal structure (e.g. No.55, No.57, and No.73), there are also series exhibiting both trending and seasonal behavior (e.g. No.1, No.11, and No.12).

As such, three datasets of 20 Hénon time series, 20 Mackey-Glass time series, and 111 NN3 time series are used for evaluating the performances of the proposed M-SVR using MIMO strategy and the counterparts in this study. Each series is split into an estimation sample and a hold-out sample. The last 18 observations are saved for evaluating and comparing the out-of-sample forecast performances of the various prediction models. All performance comparisons are based on these $18 \times 20$ out-of-sample points for Hénon and Mackey-Glass datasets and $18 \times 111$ out-of-sample points for NN3 datasets.

---

[1] The toolbox can be obtained from http://www.mathworks.com/matlabcentral/fileexchange/1597
[2] The datasets can be obtained from http://www.neural-forecasting-competition.com/NN3/datasets.htm

Normalization is a standard requirement for time series modeling and prediction. Thus, the data sets were firstly preprocessed by adopting liner transference to adjust the original data set scaled into the range of [0, 1]. Note that most of the time series in NN3 and Mackey-Glass dataset exhibit strong seasonal component or trend pattern. After the linear transference, deseasonalizing and detrending were performed. We conducted deseasonalizing by means of the revised multiplicative seasonal decomposition presented in [33]. Detrending was performed by fitting a polynomial time trend and then subtracting the estimated trend from the series when trend is detected by the Mann-Kendall test [34].

## 3.2 Accuracy measure

To compare the effectiveness of the different model, no single accuracy measure can capture the distributional features of the errors when summarized across data series. For each forecast horizon $h$, here, we consider three alternative forecast accuracy measures: the mean absolute percentage error (MAPE), symmetric mean absolute percentage error (SMAPE), and mean absolute scaled error (MASE). MAPE has the advantage of being scale-independent, and so are frequently used to compare forecast performance across different datasets. However, the MAPE also have the disadvantage that they put a heavier penalty on positive errors than on negative errors [35]. This observation led to the SMAPE which is the main measure considered in NN3 competition [36]. MASE has recently been suggested by Hyndman and Koehler [35] as a means of overcoming observation and errors around zero existing in some measures. The MASE has some features which are better than the SMAPE, which has been criticized for the fact that its treatment of positive and negative errors is not symmetric [37]. However, because of their widespread use, the MAPE and SMAPE will still be used in this study. The definitions of them are shown as follows:

$$\text{MAPE}_h = \frac{1}{S} \sum_{s=1}^{S} \left| \frac{\varphi_{t+h}^s - \hat{\varphi}_{t+h}^s}{\varphi_{t+h}^s} \right| * 100 \tag{15}$$

$$\text{SMAPE}_h = \frac{1}{S} \sum_{s=1}^{S} \frac{\left| \varphi_{t+h}^s - \hat{\varphi}_{t+h}^s \right|}{\left( \varphi_{t+h}^s + \hat{\varphi}_{t+h}^s \right) / 2} * 100 \tag{16}$$

$$\text{MASE}_h = \frac{1}{S} \sum_{s=1}^{S} \left| \frac{\varphi_{t+h}^s - \hat{\varphi}_{t+h}^s}{\frac{1}{M-1} \sum_{i=2}^{M} \left| \varphi_i^s - \varphi_{i-1}^s \right|} \right| \tag{17}$$

where $\hat{\varphi}_{t+h}^s$ is the $h$-step-ahead forecast for time series $s$, $\varphi_{t+h}^s$ is the true time series value for series $s$, $H$ is the prediction horizon (in this case $H = 18$), $S$ is the number of time series in the datasets (in this case, $S = 20$ for Hénon and Mackey-Glass datasets and $S = 111$ for NN3 datasets), and $M$ is the number of observation in the estimation sample for time series $s$. Note that these accuracy measures are computed after rolling back all of the preprocessing steps performed, such as the normalization, deseasonalizing and detrending.

## 3.3 Input selection

Filter method was employed for input selection in this study. In the case of the filter method, the best subset of inputs is selected *a priori* based only on the dataset. The input subset is chosen by a pre-defined criterion, which measures the relationship of each subset of input variables with the output. Specifically, in terms of input selection criteria, the partial mutual information (PMI) [38] was used for the ITER-SVR and DIR-SVR, while an extension of the Delta test [39] was used for the MIMO-SVR. We use PMI[3] for iterative and direct strategy because PMI is suitable for dealing with single output but not capable of dealing with multiple outputs, which leads to taking the extended Delta test as the criteria in the case of MIMO strategy, but in essence they all belong to filter methods. The maximum embedding order $d$ was set to 12 for NN3 datasets with a reference to [7] and 20 for Hénon and Mackey-Glass datasets.

## 3.4 SVR implementation

LibSVM (version 2.86) [40] and M-SVR [19-21] were employed for standard SVR and multiple-output SVR modeling in this study, respectively. We selected the Radial basis function (RBF) as the kernel function through preliminary simulation. To determine the hyper-parameters, namely $C, \varepsilon, \gamma$ (in the case of RBF as the kernel function), a population-based search algorithm, named particle swarm optimization (PSO) [41], is employed to search in the parameters space in the current study. In solving hyper-parameter selection by the PSO, each particle is requested to represent a potential solution $(C, \varepsilon, \gamma)$, namely hyper-parameters combination. Concerning the selection of parameters in PSO, it is yet another challenging model selection task [42]. Fortunately, several empirical and theoretical studies have been performed about the parameters of PSO from which valuable information can be obtained [43, 44]. In this study, the parameters are determined according to the recommendations in these studies and selected in a trial-error fashion. Table 2 below summaries the final parameters of PSO.

---

It should be noted that the input selection and parameters tuning are two independent tasks in this present study. Once the inputs are set to each time series through the filter method mentioned above, PSO is employed for parameters space searching and 5-fold cross validation is used for performance evaluation, this is, altogether PSO plus 5-fold cross validation produces the optimal parameters for SVR and MSVR models based on the training sets.

**\<Insert Table 2 here\>**

## 3.5 Experimental procedure

Fig. 1 shows the experimental procedure using the simulated and real time series. Each series is split into the estimation sample and the hold-out sample firstly. Then, the input selection and model selection for each series are conducted using aforementioned filter method, PSO algorithm, and fivefold cross-validation with iterated, direct, and MIMO strategies. Finally the attained models are tested for hold-out samples, the $\text{MAPE}_h$, $\text{SMAPE}_h$, and $\text{MASE}_h$ are computed for each prediction horizon $h$ (in our case $h=1, 2,\ldots, 18$) over three datasets (i.e., Hénon, Mackey-Glass, and NN3 datasets). Furthermore, the modeling process for each series is repeated sixty times. Upon the termination of this loop, performance of the examined models with selected strategies at each prediction horizon is judged in terms of the mean, averaged by sixty, of the $\text{MAPE}_h$, $\text{SMAPE}_h$, and $\text{MASE}_h$. Analysis of variance (ANOVA) test procedures are used to determine if the means of performance measures are statistically different among the five models for each prediction horizon and dataset. If so, Tukey's honesty significant difference (HSD) tests [45] are then employed further to identify the significantly different prediction models in multiple pair wise comparisons at 0.05 significance level.

<Insert Fig.1 here>

## 4. Results and discussion

The prediction performances of all the examined models (i.e., Naïve, Seasonal Naïve, ITER-SVR, DIR-SVR, and MIMO-SVR) in terms of three accuracy measures (i.e., MAPE, SMAPE, and MASE) and average rank for the Hénon, Mackey-Glass, and NN3 datasets are shown in Tables 3-5, respectively. The column labeled as 'Estimation sample' shows the in-sample prediction performance. The columns labeled as 'Forecast horizon $h$' show that accuracy measures at the forecast horizon $h$. The columns labeled as 'Average 1-$h$' show that average accuracy measures over the forecast horizon 1 to $h$. The last column shows the average ranking for each model over all forecast horizons of the out-of-sample prediction performance. In additions, Fig.2 depicts three representative examples, i.e., No. 55 time series in NN3 dataset, No. 11 time series in Hénon dataset, and No. 14 time series in Mackey-Glass dataset, of actual values vs. predicted values on hold-out sample.

<Insert Tables 3-5 here>

<Insert Fig.2 here>

As per the results presented, one can deduce the following observation:

- For the NN3 dataset, the top three models (according to the MAPE) turned out to be DIR-SVR, then ITER-SVR, and then MIMO-SVR. The rankings with respect to SMAPE or MASE measure are: MIMO-SVR, then DIR-SVR,

Seasonal Naïve, ITER-SVR, and Naïve.

- For the Hénon dataset, the top three models (according to the MAPE) turned out to be MIMO-SVR, then ITER-SVR, and then Naïve and Seasonal Naïve almost tie. The rankings with respect to SMAPE measure are: MIMO-SVR, then DIR-SVR, Naïve and Seasonal Naïve almost tie, and ITER-SVR. The rankings with respect to MASE measure are: MIMO-SVR, then Naïve and Seasonal Naïve almost tie, DIR-SVR, and ITER-SVR.

- For the Mackey-Glass dataset, the top three models (according to the MAPE) turned out to be MIMO-SVR, then DIR-SVR, and then ITER-SVR. The rankings with respect to SMAPE or MASE measure are: MIMO-SVR, then DIR-SVR, ITER-SVR, Seasonal Naïve, and Naïve.

- Overall, it is clear that the proposed M-SVR using MIMO strategy (i.e., MIMO-SVR) is with position within top one, although M-SVR has rarely (if ever) been considered for multi-step-ahead time series prediction in the literature. But one exception occurs when NN3 dataset is used and MAPE is considered, in which the DIR-SVR and ITER-SVR outperforms the MIMO-SVR.

- Concerning the average accuracy measures, we can see that, whatever the short ( $1 \leq h \leq 6$ ), medium ( $7 \leq h \leq 12$ ), or long ( $13 \leq h \leq 18$ ) horizon examined, whatever the dataset used, and whatever the accuracy measure considered, MIMO-SVR and DIR-SVR consistently achieve better accurate forecasts than ITER-SVR, even with a few exceptions. It is conceivable that the reason for the inferiority of ITER-SVR is that the accumulation of errors in iterated case drastically deteriorates the accuracy of the prediction. In addition, as far as the comparison between MIMO-SVR and DIR-SVR is concerned, the MIMO-SVR emerges the winner, even with a few exceptions. It is conceivable that the reason for the superiority of MIMO-SVR is that it preserves, among the predicted values, the stochastic dependency characterizing the time series.

Following [46], we also conduct a number of statistical tests to the statistical significance of any two competing models at 0.05 significance level. For each performance measure, prediction horizon, and dataset, we perform an ANOVA procedure to determine if there exists statistically significant difference among the five models in hold-out sample. The results are omitted here to save space. All ANOVA results are significant at the 0.05 level (with the exception of the horizon 6 and 18 for MAPE, horizon 6, 10, and 11 for SMAPE, and horizon 7, 8, 11, and 15 for MASE on NN3 dataset; horizon 15, 16, 17, and 18 for SMAPE on Hénon dataset), suggesting that there are significant differences among the five models. To further identify the significant difference between any two models, the Tukey's HSD test was used to compare all pairwise differences simultaneously in the current study. Note that Tukey's HSD test is a post-hoc test, this means that a researcher should not perform Tukey's HSD test unless the results of ANOVA are positive. The results of these multiple comparison tests for Hénon, Mackey-Glass, and NN3 datasets are shown in Tables 6-8, respectively. For each accuracy measure, prediction horizon, and dataset, we rank order the models from 1 (the best) to 5 (the worst).

<div align="center">**<Insert Tables 6-8 here>**</div>

Several observations can be made from Tables 6-8.

- Generally speaking, for the NN3 dataset, the difference in prediction performance among MIMO-SVR, DIR-SVR, and ITER-SVR is not significant at the 0.05 level, with some exceptions, where MIMO-SVR and ITER-SVR significantly outperform the ITER-SVR.

- When considering the Hénon dataset, the MIMO-SVR significantly outperforms the other competitors for the majority of prediction horizons. Concerning the two single-output strategies, the DIR-SVR significantly outperforms the ITER-SVR for the overwhelming majority of prediction horizons. In addition, the ITER-SVR performs the poorest at 95% statistical confidence level in most cases, particularly for MAPE and MASE measures.

- When considering the Mackey-Glass dataset, the MIMO-SVR significantly outperforms the other competitors for the majority of prediction horizons. As far as the comparison DIR-SVR vs. ITER-SVR is concerned, the difference in prediction performance is not significant at the 0.05 level, even with a few exceptions.

The computational costs of the examined models for multi-step-ahead prediction are different. From a practical viewpoint, the computational load is an important and critical issue. As the multi-step-ahead prediction may be used for optimization purposes in real-world cases, their low construction cost is a real advantage for the underlying approach. Therefore, it is reasonable to compare the examined models for their computational cost. The elapsed time of ITER-SVR, DIR-SVR, and MIMO-SVR for a single replicate for each series on Hénon, Mackey-Glass, and NN3 dataset are presented in Figs. 3-5, respectively. It is noting that the elapsed time of the Naïve and Seasonal Naïve models are negligible and thus not listed out. Details of elapsed time for ITER-SVR, DIR-SVR, and MIMO-SVR on three datasets are provided as supplements. All the numerical experiments are performed on a personal computer, Inter(R) Core(TM) 2 Duo CPU 2.50 GHz, 1.87-GB memory, and MATLAB environment (Version R2009b).

**<Insert Figs. 3-5 here>**

According to the obtained results, one can deduce the following observations:

- The DIR-SVR is computationally much more expensive than the ITER-SVR and MIMO-SVR. Both ITER-SVR and MIMO-SVR are tens of times faster than the DIR-SVR across three datasets.

- The elapsed time of iterated strategy and MIMO strategy increase slightly with the prediction horizon. However, the computational cost of DIR-SVR increases drastically with the sample size of series.

- The ITER-SVR is the least expensive model, but the difference of computational cost between ITER-SVR and MIMO-SVR is negligible, particularly for the small sample case.

## 5. Conclusions

As a well-known intelligent algorithm, SVR is a well-established and well-tested technique for multi-step-ahead prediction. However, the standard formulation of SVR using conventional prediction strategies for multi-step-ahead prediction suffer from either from error accumulation, like in the ITER-SVR, or from expensive computational cost, like in the DIR-SVR. This paper assessed the performance of the novel application of multiple-output SVR (M-SVR) using MIMO strategy for multi-step-ahead time series prediction, and then goes a step forward by comparatively examined the rank of three leading prediction strategies with SVR. Specifically, quantitative and comprehensive assessments were performed with the simulated and real datasets on the basis of the prediction accuracy and computational cost. In addition, Naïve and Seasonal Naïve are selected as benchmarks. According to the obtained results, the MIMO-SVR achieved consistently better prediction performance than DIR-SVR and ITER-SVR in terms of MAPE, SMAPE, and MASE across three datasets, even with a few exceptions. However, the difference of prediction performance between DIR-SVR and ITER-SVR is not significant at the 0.05 level in the most cases. The computational load of DIR-SVR is extremely expensive compared to the two competitors. However, the difference of computational load between ITER-SVR and MIMO-SVR is negligible. Results indicate that the MIMO-SVR is a very promising technique with high-quality forecasts and accredited computational loads for multi-step-ahead time series prediction.

The limitations of this study lie in two aspects. First, we have used only SVR as the modeling technique; future research could examine more modeling technique, such as neural networks, to substantiate our findings. Second, our experimental study focuses on three commonly used prediction strategies. Further research is needed to investigate the performance of multi-step-ahead time series prediction with richer strategies.

## Acknowledgment

This work was supported by Natural Science Foundation of China under Project Nos. 70771042 and the Fundamental Research Funds for the Central Universities (2012QN208-HUST) and a grant from the Modern Information Management Research Center at Huazhong University of Science and Technology.

**Figures**

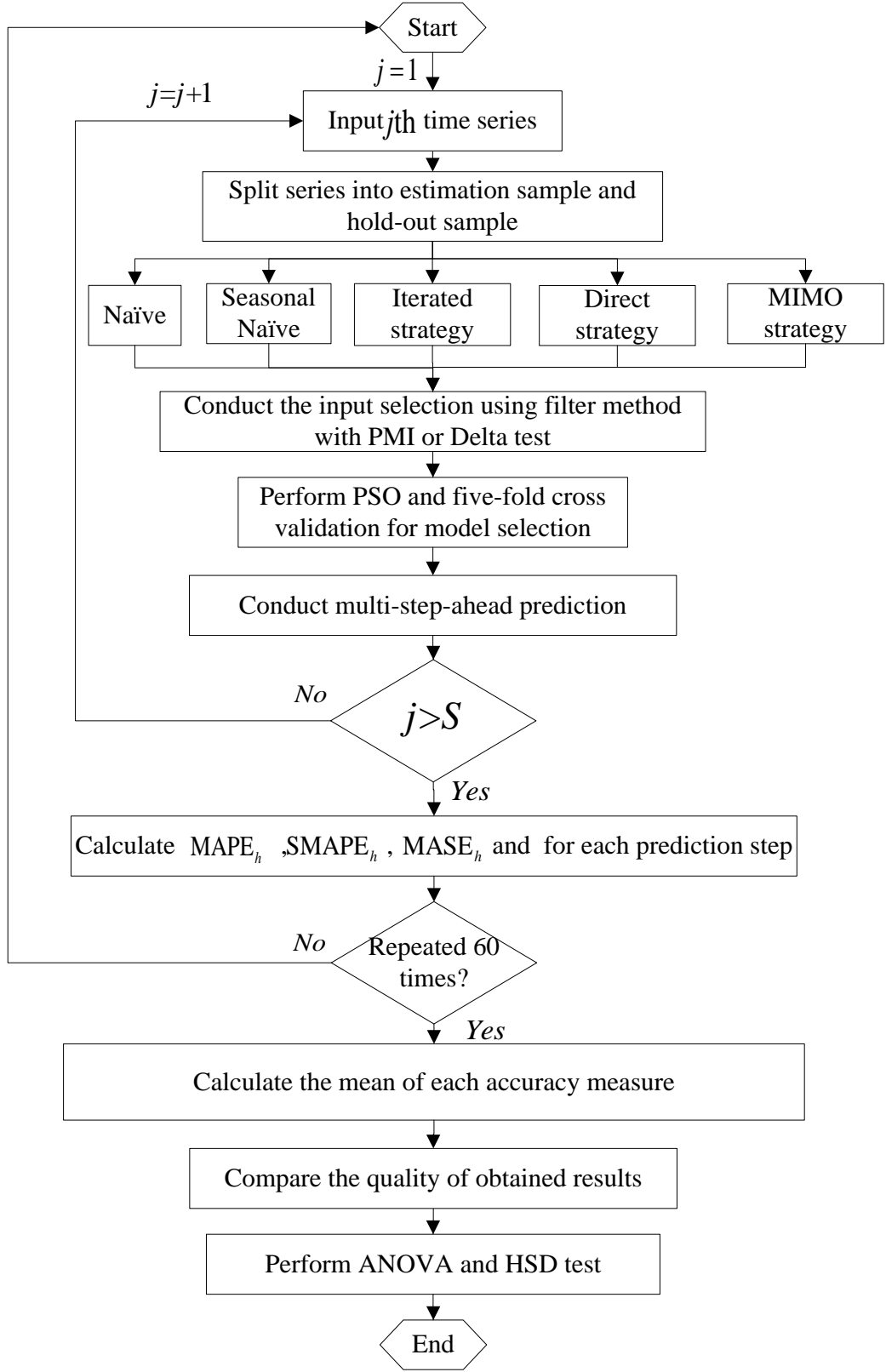

Fig. 1 Experiment procedure for multi-step-ahead prediction

(a)

(b)

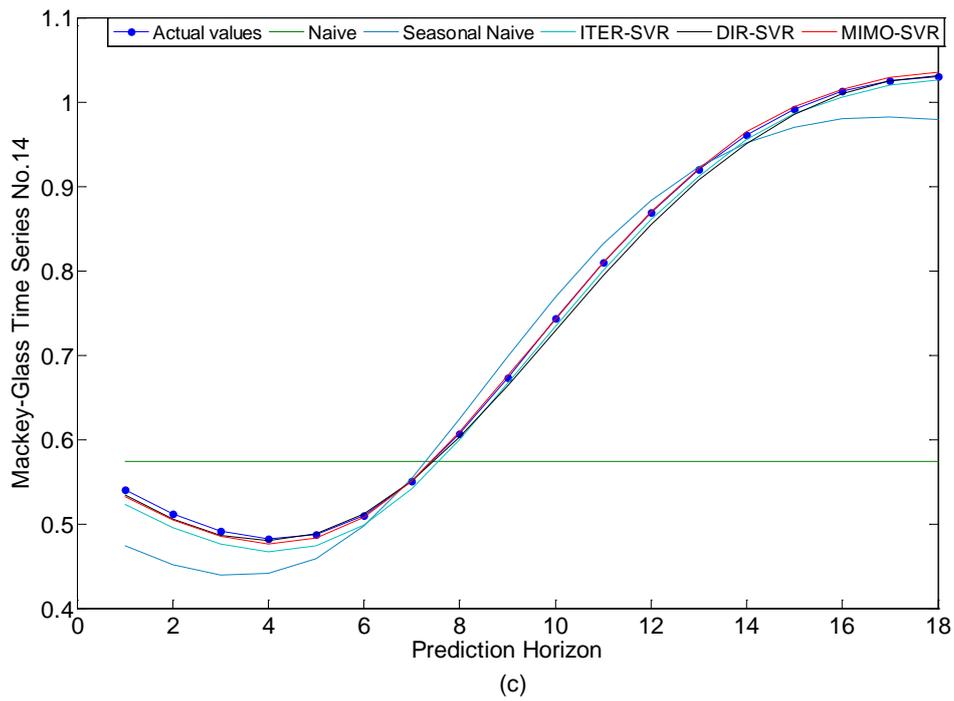

Fig. 2 Three representative examples of actual values vs. predicted values on hold-out sample: (a) No. 55 time series in NN3 dataset, (b) No. 11 time series in Hénon dataset, and (c) No. 14 time series in Mackey-Glass dataset.

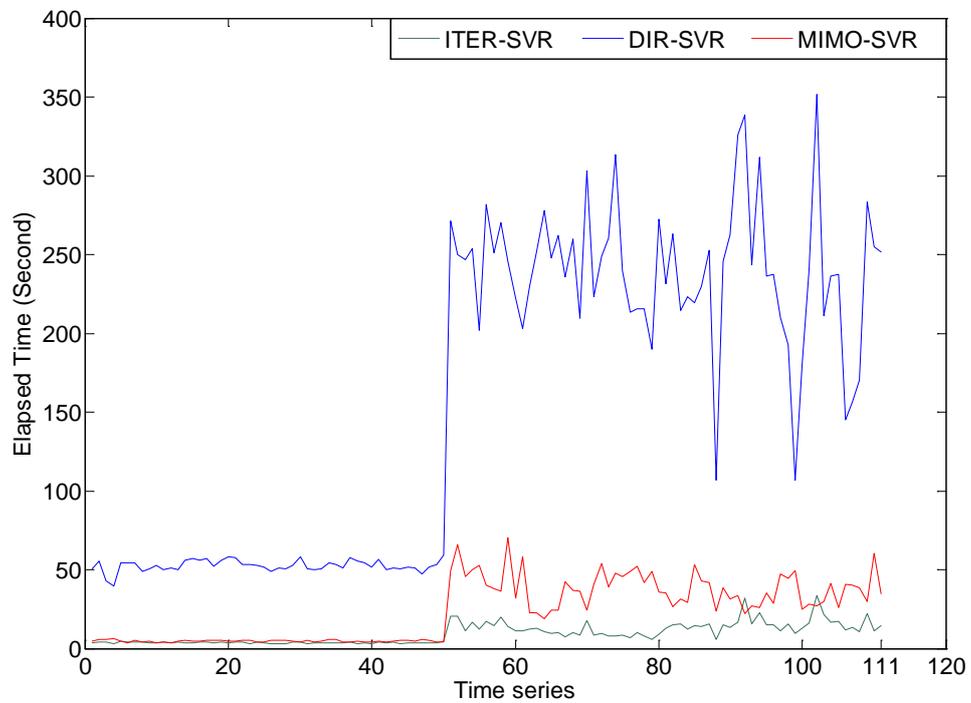

Fig. 3 Elapsed time of three models for each series of NN3 dataset

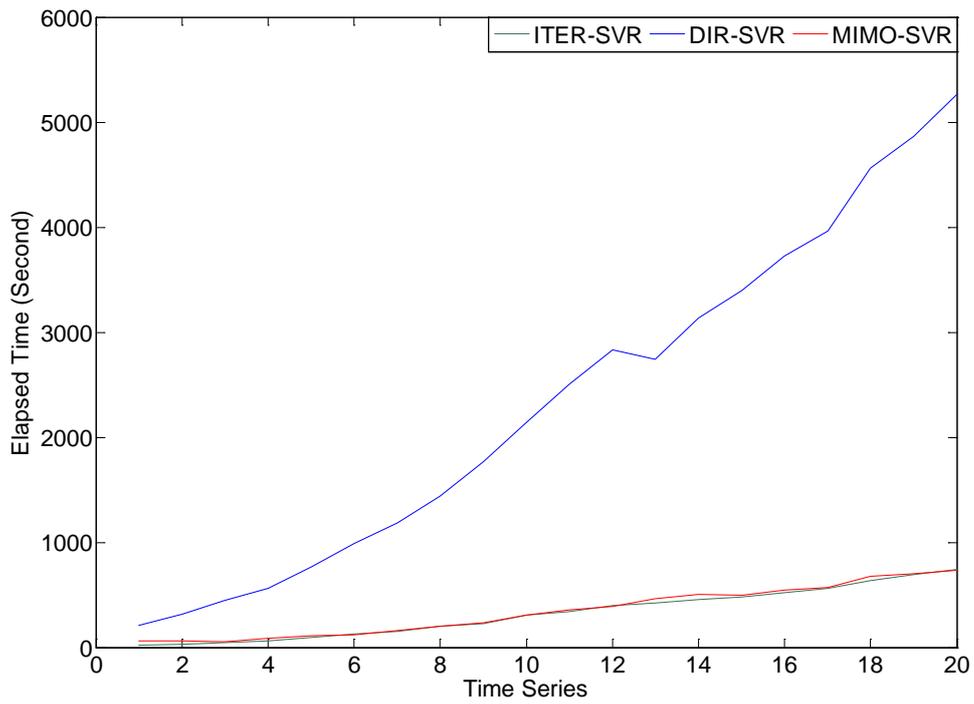

Fig. 4 Elapsed time of three models for each series of Hénon dataset

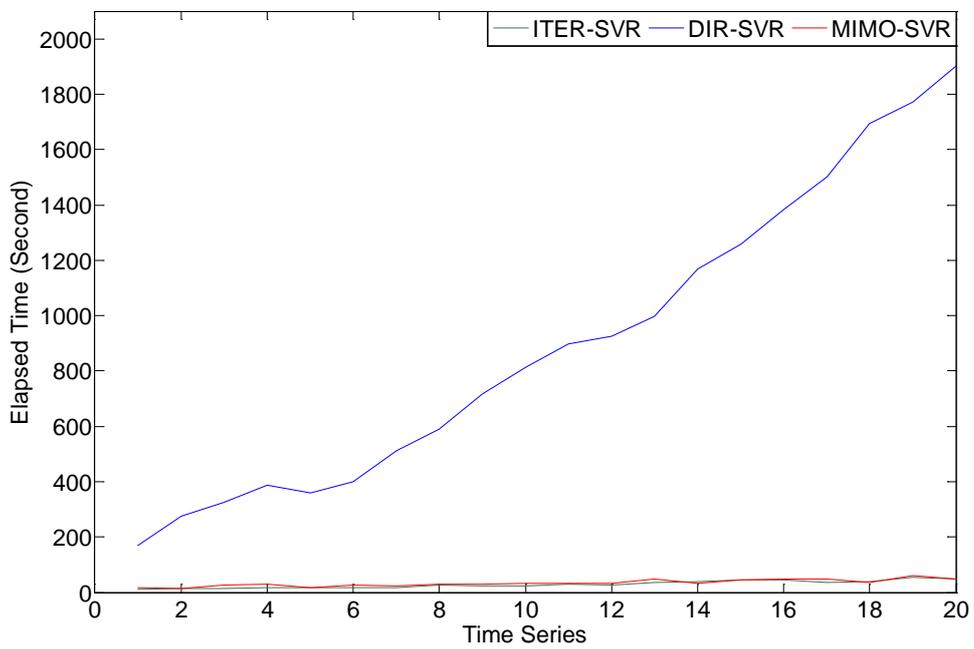

Fig. 5 Elapsed time of three models for each series of Mackey-Glass dataset

# Tables

Table 1 Initialization and sample size of the simulated time series

| No. | DGP | | | | Sample size |
|---|---|---|---|---|---|
| | Hénon ( $\phi_1, \varphi_1$ ) | | Mackey-Glass ( $\varphi_1, \tau$ ) | | |
| 1 | 0.100 | 0.100 | 1.000 | 15 | 205 |
| 2 | 0.100 | 0.300 | 1.200 | 15 | 246 |
| 3 | 0.100 | 0.500 | 1.400 | 15 | 297 |
| 4 | 0.100 | 0.700 | 1.600 | 15 | 341 |
| 5 | 0.100 | 0.900 | 1.800 | 15 | 389 |
| 6 | 0.300 | 0.100 | 2.000 | 15 | 428 |
| 7 | 0.300 | 0.300 | 1.000 | 16 | 489 |
| 8 | 0.300 | 0.500 | 1.200 | 16 | 534 |
| 9 | 0.300 | 0.700 | 1.400 | 16 | 584 |
| 10 | 0.300 | 0.900 | 1.600 | 16 | 648 |
| 11 | 0.500 | 0.100 | 1.800 | 16 | 685 |
| 12 | 0.500 | 0.300 | 2.000 | 16 | 718 |
| 13 | 0.500 | 0.500 | 1.000 | 17 | 745 |
| 14 | 0.500 | 0.700 | 1.200 | 17 | 784 |
| 15 | 0.500 | 0.900 | 1.400 | 17 | 804 |
| 16 | 0.700 | 0.100 | 1.600 | 17 | 834 |
| 17 | 0.700 | 0.300 | 1.800 | 17 | 879 |
| 18 | 0.700 | 0.500 | 2.000 | 17 | 915 |
| 19 | 0.700 | 0.700 | 1.000 | 18 | 957 |
| 20 | 0.700 | 0.900 | 1.200 | 18 | 986 |

Table 2 Parameter selection of the PSO

| Parameters | Values |
|---|---|
| Swarm size | 20 |
| Number of iterations | 100 |
| Cognitive coefficients | 2.0 |
| Interaction coefficients | 2.0 |
| Initial weight | 0.9 |
| Final weight | 0.4 |

Table 3 Prediction accuracy measures for NN3 dataset

| Model | Estimation sample | Hold-out sample | | | | | | | | | | Average rank |
|---|---|---|---|---|---|---|---|---|---|---|---|---|
| | | Forecast horizon($h$) | | | | | | Average 1-$h$ | | | | |
| | | 1 | 2 | 3 | 6 | 12 | 18 | 1-6 | 7-12 | 13-18 | 1-18 | |
| | MAPE | | | | | | | | | | | |
| Naïve | 34.944 | 25.786 | 36.972 | 31.427 | 24.682 | 23.190 | 24.799 | 29.775 | 28.192 | 33.108 | 30.359 | 4.500 |
| Seasonal Naïve | 26.418 | 24.418 | 28.533 | 24.624 | 19.310 | 23.190 | 22.672 | 23.553 | 24.096 | 26.470 | 24.707 | 3.333 |
| ITER-SVR | 22.481 | 11.541 | 8.818 | 10.464 | 21.665 | 25.232 | 21.200 | 13.772 | 30.459 | 32.380 | 25.537 | 2.556 |
| DIR-SVR | 18.105 | 11.577 | 8.886 | 10.543 | 23.472 | 17.604 | 22.428 | 13.400 | 19.758 | 25.578 | 19.579 | 1.944 |
| MIMO-SVR | 18.248 | 11.649 | 9.052 | 10.633 | 17.320 | 28.694 | 24.885 | 12.678 | 23.060 | 23.480 | 19.739 | 2.667 |
| | SMAPE | | | | | | | | | | | |
| Naïve | 22.823 | 19.512 | 19.439 | 22.812 | 23.014 | 19.390 | 25.886 | 22.127 | 21.392 | 24.145 | 22.554 | 4.722 |
| Seasonal Naïve | 17.824 | 16.969 | 17.238 | 20.805 | 16.629 | 19.390 | 22.277 | 17.793 | 17.959 | 19.783 | 18.512 | 2.667 |
| ITER-SVR | 16.895 | 11.229 | 8.824 | 10.916 | 19.879 | 22.897 | 22.409 | 13.571 | 20.036 | 21.872 | 18.493 | 3.278 |
| DIR-SVR | 15.421 | 11.209 | 8.909 | 11.031 | 20.724 | 18.445 | 21.781 | 12.962 | 18.836 | 19.782 | 17.193 | 2.333 |
| MIMO-SVR | 13.815 | 11.262 | 9.077 | 11.074 | 17.370 | 19.153 | 21.640 | 12.487 | 18.505 | 18.984 | 16.659 | 2.000 |
| | MASE | | | | | | | | | | | |
| Naïve | 1.338 | 1.003 | 1.050 | 1.205 | 1.497 | 1.185 | 2.153 | 1.236 | 1.450 | 1.752 | 1.479 | 4.167 |
| Seasonal Naïve | 1.315 | 1.058 | 1.123 | 1.249 | 1.197 | 1.185 | 1.915 | 1.138 | 1.298 | 1.537 | 1.324 | 3.000 |
| ITER-SVR | 1.321 | 0.485 | 0.464 | 0.582 | 1.529 | 1.619 | 1.905 | 0.855 | 1.504 | 1.664 | 1.341 | 3.444 |
| DIR-SVR | 1.174 | 0.491 | 0.472 | 0.604 | 1.399 | 1.417 | 1.582 | 0.769 | 1.453 | 1.393 | 1.205 | 2.667 |
| MIMO-SVR | 0.875 | 0.511 | 0.497 | 0.585 | 1.209 | 1.380 | 1.617 | 0.705 | 1.222 | 1.325 | 1.084 | 1.722 |

Table 4 Prediction accuracy measures for Hénon dataset

| Model | Estimation sample | Hold-out sample | | | | | | | | | | Average rank |
|---|---|---|---|---|---|---|---|---|---|---|---|---|
| | | Forecast horizon($h$) | | | | | | Average 1-$h$ | | | | |
| | | 1 | 2 | 3 | 6 | 12 | 18 | 1-6 | 7-12 | 13-18 | 1-18 | |
| | MAPE | | | | | | | | | | | |
| Naïve | 171.636 | 260.851 | 198.318 | 217.091 | 264.451 | 239.739 | 199.046 | 222.840 | 193.433 | 196.262 | 204.178 | 2.889 |
| Seasonal Naïve | 171.636 | 260.851 | 198.318 | 217.091 | 264.451 | 239.739 | 199.046 | 222.840 | 193.433 | 196.262 | 204.178 | 2.889 |
| ITER-SVR | 251.845 | 72.058 | 1035.228 | 1368.830 | 308.521 | 193.779 | 349.631 | 550.988 | 151.576 | 191.283 | 297.949 | 2.611 |
| DIR-SVR | 205.184 | 40.077 | 449.664 | 144.864 | 351.826 | 232.521 | 325.959 | 242.045 | 217.288 | 260.037 | 239.790 | 3.556 |
| MIMO-SVR | 210.547 | 25.036 | 127.897 | 105.528 | 209.935 | 145.697 | 389.448 | 114.532 | 335.317 | 273.885 | 241.245 | 2.500 |
| | SMAPE | | | | | | | | | | | |
| Naïve | 118.483 | 136.654 | 119.228 | 157.677 | 162.404 | 170.906 | 166.033 | 148.540 | 136.183 | 136.595 | 140.439 | 3.167 |
| Seasonal Naïve | 118.483 | 136.654 | 119.228 | 157.677 | 162.404 | 170.906 | 166.033 | 148.540 | 136.183 | 136.595 | 140.439 | 3.167 |
| ITER-SVR | 115.841 | 58.928 | 118.636 | 113.307 | 120.217 | 159.064 | 160.939 | 114.449 | 147.074 | 146.693 | 136.072 | 3.500 |
| DIR-SVR | 109.512 | 46.485 | 119.885 | 104.423 | 152.817 | 127.639 | 151.030 | 106.975 | 131.836 | 137.304 | 125.371 | 2.667 |
| MIMO-SVR | 101.956 | 17.992 | 53.458 | 59.789 | 132.865 | 121.563 | 148.239 | 68.397 | 116.117 | 129.110 | 125.371 | 1.500 |
| | MASE | | | | | | | | | | | |
| Naïve | 0.784 | 0.990 | 0.796 | 1.079 | 1.093 | 1.176 | 1.199 | 0.999 | 0.896 | 0.911 | 0.935 | 2.444 |
| Seasonal Naïve | 0.784 | 0.990 | 0.796 | 1.079 | 1.093 | 1.176 | 1.199 | 0.999 | 0.896 | 0.911 | 0.935 | 2.444 |
| ITER-SVR | 1.524 | 0.690 | 0.908 | 0.889 | 1.292 | 2.286 | 2.210 | 1.125 | 1.899 | 1.874 | 1.633 | 4.667 |
| DIR-SVR | 0.742 | 0.597 | 0.871 | 0.873 | 1.118 | 1.085 | 1.066 | 0.854 | 1.020 | 1.015 | 0.963 | 3.000 |
| MIMO-SVR | 0.548 | 0.088 | 0.207 | 0.271 | 0.813 | 0.874 | 0.955 | 0.359 | 0.739 | 0.878 | 0.659 | 1.500 |

Table 5 Prediction accuracy measures for Mackey-Glass dataset

| Model | Estimation sample | Hold-out sample | | | | | | | | | | Average rank |
|---|---|---|---|---|---|---|---|---|---|---|---|---|
| | | Forecast horizon($h$) | | | | | | Average 1-$h$ | | | | |
| | | 1 | 2 | 3 | 6 | 12 | 18 | 1-6 | 7-12 | 13-18 | 1-18 | |
| | MAPE | | | | | | | | | | | |
| Naïve | 21.932 | 3.289 | 6.702 | 10.012 | 18.857 | 40.549 | 46.179 | 11.267 | 31.613 | 45.444 | 29.441 | 4.889 |
| Seasonal Naïve | 6.482 | 7.660 | 7.068 | 6.178 | 9.003 | 7.519 | 9.868 | 7.481 | 8.691 | 9.481 | 8.551 | 4.111 |
| ITER-SVR | 0.821 | 1.018 | 1.016 | 1.004 | 1.034 | 0.819 | 1.001 | 1.047 | 0.785 | 0.908 | 0.913 | 2.778 |
| DIR-SVR | 0.813 | 0.701 | 0.909 | 1.001 | 0.977 | 0.765 | 1.000 | 0.962 | 0.827 | 0.861 | 0.883 | 2.167 |
| MIMO-SVR | 0.598 | 0.748 | 0.709 | 0.697 | 0.625 | 0.628 | 0.694 | 0.710 | 0.617 | 0.623 | 0.650 | 1.056 |
| | SMAPE | | | | | | | | | | | |
| Naïve | 118.483 | 3.332 | 6.834 | 10.205 | 18.833 | 38.718 | 43.688 | 11.374 | 30.697 | 42.958 | 28.343 | 4.889 |
| Seasonal Naïve | 6.084 | 7.815 | 7.152 | 6.276 | 8.594 | 7.317 | 9.582 | 7.464 | 8.175 | 9.256 | 8.298 | 4.111 |
| ITER-SVR | 0.921 | 1.017 | 1.014 | 1.005 | 1.037 | 0.822 | 1.002 | 1.048 | 0.787 | 0.910 | 0.915 | 2.722 |
| DIR-SVR | 0.841 | 0.703 | 0.912 | 1.005 | 0.978 | 0.764 | 0.998 | 0.965 | 0.824 | 0.860 | 0.883 | 2.222 |
| MIMO-SVR | 0.524 | 0.749 | 0.709 | 0.698 | 0.628 | 0.629 | 0.694 | 0.712 | 0.617 | 0.623 | 0.651 | 1.056 |
| | MASE | | | | | | | | | | | |
| Naïve | 6.328 | 0.925 | 1.888 | 2.828 | 5.271 | 10.857 | 12.811 | 3.176 | 8.546 | 12.442 | 8.055 | 4.889 |
| Seasonal Naïve | 1.845 | 2.080 | 1.964 | 1.797 | 2.284 | 2.028 | 2.417 | 2.032 | 2.229 | 2.298 | 2.186 | 4.111 |
| ITER-SVR | 0.215 | 0.286 | 0.272 | 0.273 | 0.284 | 0.228 | 0.278 | 0.286 | 0.217 | 0.253 | 0.252 | 2.667 |
| DIR-SVR | 0.198 | 0.194 | 0.250 | 0.281 | 0.257 | 0.208 | 0.286 | 0.263 | 0.217 | 0.241 | 0.240 | 2.278 |
| MIMO-SVR | 0.154 | 0.203 | 0.186 | 0.195 | 0.178 | 0.172 | 0.192 | 0.195 | 0.166 | 0.175 | 0.179 | 1.056 |

Table 6 Multiple comparison results with ranked models for hold-out sample on NN3 dataset

| Measure | Prediction Horizon | 1 | | 2 | | 3 | | 4 | | 5 |
|---|---|---|---|---|---|---|---|---|---|---|
| | | 1 | | 2 | | 3 | | 4 | | 5 |
| MAPE$_h$ | 1-3,7 | ITER-SVR | < | DIR-SVR | < | MIMO-SVR | <* | S-Naïve | < | Naive |
| | 4 | DIR-SVR | < | ITER-SVR | < | MIMO-SVR | <* | S-Naive | < | Naïve |
| | 5 | DIR-SVR | < | MIMO-SVR | <* | ITER-SVR | < | S-Naive | <* | Naïve |
| | 8 | MIMO-SVR | < | DIR-SVR | < | S-Naïve | <* | Naive | <* | ITER-SVR |
| | 9 | MIMO-SVR | < | DIR-SVR | < | ITER-SVR | <* | S-Naive | <* | Naive |
| | 10 | ITER-SVR | < | DIR-SVR | < | S-Naive | <* | S-Naive | < | MIMO-SVR |
| | 11 | DIR-SVR | < | ITER-SVR | < | MIMO-SVR | <* | S-Naive | <* | Naïve |
| | 12 | DIR-SVR | <* | Naïve | < | S-Naive | < | ITER-SVR | < | MIMO-SVR |
| | 13 | DIR-SVR | <* | ITER-SVR | < | MIMO-SVR | <* | S-Naive | < | Naïve |
| | 14 | MIMO-SVR | < | DIR-SVR | <* | S-Naïve | <* | Naive | <* | ITER-SVR |
| | 15 | MIMO-SVR | < | S-Naïve | < | DIR-SVR | < | ITER-SVR | <* | Naive |
| | 16 | DIR-SVR | < | ITER-SVR | <* | S-Naïve | < | MIMO-SVR | <* | Naive |
| | 17 | MIMO-SVR | <* | S-Naïve | <* | Naïve | <* | DIR-SVR | <* | ITER-SVR |
| SMAPE$_h$ | 1 | DIR-SVR | < | ITER-SVR | < | MIMO-SVR | <* | S-Naive | <* | Naive |
| | 2,3 | ITER-SVR | < | DIR-SVR | < | MIMO-SVR | <* | S-Naive | < | Naive |
| | 4 | ITER-SVR | < | DIR-SVR | < | MIMO-SVR | <* | S-Naive | <* | Naïve |
| | 5 | DIR-SVR | < | MIMO-SVR | <* | S-Naive | < | ITER-SVR | <* | Naive |
| | 7 | ITER-SVR | < | MIMO-SVR | < | DIR-SVR | < | S-Naive | <* | Naive |
| | 8 | S-Naive | < | MIMO-SVR | < | DIR-SVR | <* | ITER-SVR | < | Naive |
| | 9 | MIMO-SVR | < | S-Naive | < | DIR-SVR | < | ITER-SVR | <* | Naive |
| | 12 | DIR-SVR | < | MIMO-SVR | < | Naive | < | S-Naive | <* | ITER-SVR |
| | 13 | DIR-SVR | < | S-Naive | < | MIMO-SVR | <* | ITER-SVR | < | Naïve |
| | 14 | MIMO-SVR | < | S-Naive | < | DIR-SVR | <* | ITER-SVR | < | ITER-SVR |
| | 15, 16,18 | MIMO-SVR | < | DIR-SVR | < | S-Naive | < | ITER-SVR | <* | Naive |
| | 17 | MIMO-SVR | < | S-Naive | < | DIR-SVR | < | Naive | <* | ITER-SVR |
| MASE$_h$ | 1,2 | ITER-SVR | < | DIR-SVR | < | MIMO-SVR | <* | Naive | < | S-Naive |
| | 3 | ITER-SVR | < | MIMO-SVR | < | DIR-SVR | <* | Naive | < | S-Naive |
| | 4 | MIMO-SVR | < | ITER-SVR | < | DIR-SVR | <* | S-Naive | < | Naive |
| | 5 | MIMO-SVR | < | DIR-SVR | < | S-Naive | < | Naive | <* | ITER-SVR |
| | 6 | S-Naive | < | MIMO-SVR | < | DIR-SVR | < | Naive | <* | ITER-SVR |
| | 9 | MIMO-SVR | <* | S-Naive | < | DIR-SVR | < | Naive | <* | ITER-SVR |
| | 10 | MIMO-SVR | <* | S-Naive | < | Naive | < | ITER-SVR | < | DIR-SVR |
| | 12 | Naive | < | S-Naive | < | MIMO-SVR | <* | DIR-SVR | <* | ITER-SVR |
| | 13 | DIR-SVR | < | S-Naive | < | MIMO-SVR | <* | ITER-SVR | <* | Naive |
| | 14 | MIMO-SVR | <* | DIR-SVR | < | S-Naive | < | Naive | <* | ITER-SVR |
| | 16 | MIMO-SVR | < | DIR-SVR | < | S-Naive | < | ITER-SVR | <* | Naive |
| | 17 | MIMO-SVR | < | DIR-SVR | <* | ITER-SVR | < | S-Naive | < | Naïve |
| | 18 | DIR-SVR | < | MIMO-SVR | <* | ITER-SVR | < | S-Naive | < | Naive |

* indicates the mean difference between the two adjacent methods is significant at the 0.05 level

Table 7 Multiple comparison results with ranked models for hold-out sample on Hénon dataset

| Measure | Prediction Horizon | Rank of Models | | | | | | | | | |
|---|---|---|---|---|---|---|---|---|---|---|---|
| | | 1 | | 2 | | 3 | | 4 | | 5 | |
| MAPE$_h$ | 1 | MIMO-SVR | <* | DIR-SVR | <* | ITER-SVR | <* | Naive | = | S-Naive | |
| | 2 | MIMO-SVR | <* | Naive | = | S-Naive | <* | DIR-SVR | <* | ITER-SVR | |
| | 3 | MIMO-SVR | <* | DIR-SVR | <* | Naive | = | S-Naive | <* | ITER-SVR | |
| | 4 | MIMO-SVR | < | DIR-SVR | <* | Naive | = | S-Naive | <* | ITER-SVR | |
| | 5 | MIMO-SVR | <* | ITER-SVR | <* | Naive | = | S-Naive | <* | DIR-SVR | |
| | 6 | MIMO-SVR | <* | Naive | = | S-Naive | <* | ITER-SVR | < | MIMO-SVR | |
| | 7 | ITER-SVR | < | DIR-SVR | < | Naïve | = | S-Naive | <* | MIMO-SVR | |
| | 8 | ITER-SVR | <* | MIMO-SVR | < | Naïve | = | S-Naive | < | DIR-SVR | |
| | 9 | Naive | = | S-Naive | <* | ITER-SVR | <* | DIR-SVR | <* | MIMO-SVR | |
| | 10 | ITER-SVR | <* | MIMO-SVR | < | Naïve | = | S-Naive | <* | DIR-SVR | |
| | 11 | ITER-SVR | <* | MIMO-SVR | <* | DIR-SVR | < | Naïve | = | S-Naive | |
| | 12 | MIMO-SVR | <* | ITER-SVR | <* | DIR-SVR | < | Naive | = | S-Naive | |
| | 13 | Naive | < | S-Naive | < | ITER-SVR | <* | DIR-SVR | <* | MIMO-SVR | |
| | 14 | ITER-SVR | <* | MIMO-SVR | < | Naïve | = | S-Naive | <* | DIR-SVR | |
| | 15 | ITER-SVR | < | MIMO-SVR | < | Naïve | = | S-Naive | <* | DIR-SVR | |
| | 16 | Naive | = | S-Naive | < | DIR-SVR | < | ITER-SVR | <* | MIMO-SVR | |
| | 17 | ITER-SVR | <* | DIR-SVR | < | MIMO-SVR | < | Naïve | = | S-Naive | |
| | 18 | Naive | < | S-Naive | <* | DIR-SVR | < | ITER-SVR | < | MIMO-SVR | |
| SMAPE$_h$ | 1,3,4 | MIMO-SVR | <* | DIR-SVR | < | ITER-SVR | <* | Naive | = | S-Naive | |
| | 2 | MIMO-SVR | <* | ITER-SVR | < | Naive | = | S-Naive | < | DIR-SVR | |
| | 5 | MIMO-SVR | <* | DIR-SVR | <* | Naive | = | S-Naive | < | ITER-SVR | |
| | 6 | ITER-SVR | < | MIMO-SVR | <* | DIR-SVR | < | Naïve | = | S-Naive | |
| | 7 | DIR-SVR | < | MIMO-SVR | <* | Naive | = | S-Naive | < | ITER-SVR | |
| | 8 | MIMO-SVR | <* | Naive | = | S-Naive | < | DIR-SVR | < | ITER-SVR | |
| | 9 | Naive | = | S-Naive | <* | MIMO-SVR | < | DIR-SVR | <* | ITER-SVR | |
| | 10 | MIMO-SVR | <* | DIR-SVR | < | Naive | = | S-Naive | < | ITER-SVR | |
| | 11 | MIMO-SVR | <* | DIR-SVR | < | ITER-SVR | < | Naive | = | S-Naive | |
| | 12 | MIMO-SVR | < | DIR-SVR | <* | ITER-SVR | < | Naive | = | S-Naive | |
| | 13 | MIMO-SVR | <* | Naive | = | S-Naive | <* | DIR-SVR | < | ITER-SVR | |
| | 14 | MIMO-SVR | < | Naive | = | S-Naive | <* | ITER-SVR | < | DIR-SVR | |
| MASE$_h$ | 1,4 | MIMO-SVR | <* | DIR-SVR | < | ITER-SVR | <* | Naive | = | S-Naive | |
| | 2,6 | MIMO-SVR | <* | Naive | = | S-Naive | < | DIR-SVR | < | ITER-SVR | |
| | 3 | MIMO-SVR | <* | DIR-SVR | < | ITER-SVR | < | Naive | = | S-Naive | |
| | 5 | MIMO-SVR | <* | DIR-SVR | < | Naive | = | S-Naive | <* | ITER-SVR | |
| | 7,11,12,16,18 | MIMO-SVR | < | DIR-SVR | < | Naive | = | S-Naive | <* | ITER-SVR | |
| | 8 | MIMO-SVR | < | Naive | = | S-Naive | <* | DIR-SVR | <* | ITER-SVR | |
| | 9,14 | Naive | = | S-Naive | < | MIMO-SVR | <* | DIR-SVR | <* | ITER-SVR | |
| | 10,13 | MIMO-SVR | <* | Naive | = | S-Naive | < | DIR-SVR | <* | ITER-SVR | |
| | 14 | Naive | = | S-Naive | < | MIMO-SVR | <* | DIR-SVR | <* | ITER-SVR | |
| | 15 | Naive | = | S-Naive | < | MIMO-SVR | < | DIR-SVR | <* | ITER-SVR | |
| | 17 | Naive | = | S-Naive | < | DIR-SVR | < | MIMO-SVR | <* | ITER-SVR | |

* indicates the mean difference between the two adjacent methods is significant at the 0.05 level

Table 8 Multiple comparison results with ranked models for hold-out sample on Mackey-Glass dataset

| Measure | Prediction Horizon | Rank of Models | | | | | | | | |
|---|---|---|---|---|---|---|---|---|---|---|
| | | 1 | | 2 | | 3 | | 4 | | 5 |
| MAPE$_h$ | 1 | DIR-SVR | < | MIMO-SVR | <* | ITER-SVR | <* | Naive | < | S-Naive |
| | 2 | MIMO-SVR | < | DIR-SVR | < | ITER-SVR | <* | Naive | < | S-Naive |
| | 3-6,13,14,16-18 | MIMO-SVR | <* | DIR-SVR | < | ITER-SVR | <* | S-Naive | <* | Naive |
| | 7 | MIMO-SVR | < | DIR-SVR | <* | ITER-SVR | <* | S-Naive | <* | Naive |
| | 8,12 | MIMO-SVR | < | DIR-SVR | < | ITER-SVR | <* | S-Naive | <* | Naive |
| | 9 | MIMO-SVR | < | ITER-SVR | <* | DIR-SVR | <* | S-Naive | <* | Naive |
| | 10,11 | MIMO-SVR | < | ITER-SVR | < | DIR-SVR | <* | S-Naive | <* | Naive |
| | 15 | MIMO-SVR | <* | ITER-SVR | < | DIR-SVR | <* | S-Naive | <* | Naive |
| SMAPE$_h$ | 1 | DIR-SVR | < | MIMO-SVR | <* | ITER-SVR | <* | Naive | < | S-Naive |
| | 2 | MIMO-SVR | <* | DIR-SVR | < | ITER-SVR | <* | Naive | < | S-Naive |
| | 3 | MIMO-SVR | <* | ITER-SVR | < | DIR-SVR | <* | S-Naive | <* | Naive |
| | 4-6,13,14,17,18 | MIMO-SVR | <* | DIR-SVR | < | ITER-SVR | <* | S-Naive | <* | Naive |
| | 7 | MIMO-SVR | < | DIR-SVR | <* | ITER-SVR | <* | S-Naive | <* | Naive |
| | 8,12,16 | MIMO-SVR | < | DIR-SVR | < | ITER-SVR | <* | S-Naive | <* | Naive |
| | 9 | MIMO-SVR | < | ITER-SVR | <* | DIR-SVR | <* | S-Naive | <* | Naive |
| | 10,11,15 | MIMO-SVR | <* | ITER-SVR | < | DIR-SVR | <* | S-Naive | <* | Naive |
| MASE$_h$ | 1 | DIR-SVR | < | MIMO-SVR | < | ITER-SVR | <* | Naive | <* | S-Naive |
| | 2,5 | MIMO-SVR | <* | DIR-SVR | < | ITER-SVR | <* | Naive | <* | S-Naive |
| | 3,9-11,18 | MIMO-SVR | < | ITER-SVR | < | DIR-SVR | <* | Naive | <* | S-Naive |
| | 4 | MIMO-SVR | <* | ITER-SVR | < | DIR-SVR | <* | Naive | <* | S-Naive |
| | 6-8,12-17 | MIMO-SVR | < | DIR-SVR | < | ITER-SVR | <* | Naive | <* | S-Naive |

* indicates the mean difference between the two adjacent methods is significant at the 0.05 level